\documentclass[11pt]{article}
\usepackage{amssymb}

\usepackage[final]{acl}
\usepackage{times}
\usepackage{latexsym}

\usepackage[T1]{fontenc}

\usepackage[utf8]{inputenc}

\usepackage{microtype}

\usepackage{inconsolata}

\usepackage{graphicx}

\usepackage{listings}
\usepackage[most]{tcolorbox}
\usepackage{listings}

\usepackage{amsmath}
\usepackage{amssymb}
\usepackage[ruled,vlined,linesnumbered]{algorithm2e}

\usepackage{booktabs}
\usepackage{makecell}
\usepackage{multirow}
\usepackage{graphicx} 

\usepackage{float}

\usepackage{cuted} 
\usepackage{capt-of}   
\usepackage{needspace} 
\usepackage{placeins}

\title{ProUIE: A Macro-to-Micro Progressive Learning Method for LLM-based Universal Information Extraction}

\author{
\textbf{Wenda Liu}$^{1}$\thanks{\ \ Equal contribution.},
\textbf{Zhigang Song}$^{1}$\footnotemark[1],
\textbf{Shuai Nie}$^{1}$\thanks{\ \ Corresponding author.},
\textbf{Guangyao Liu}$^{1}$,
\textbf{Lisung Chen}$^{1}$\\
\textbf{Binyu Yang}$^{1}$,
\textbf{Yaran Chen}$^{2}$,
\textbf{Peng Zhou}$^{1}$,
\textbf{Hongzhen Wang}$^{1}$,
\textbf{Yuchen Liu}$^{1}$\\
\textbf{Wenyue Hu}$^{1}$,
\textbf{Jiaming Xu}$^{1}$,
\textbf{Runyu Shi}$^{1}$,
\textbf{Ying Huang}$^{1}$\\
$^1$Xiaomi Corporation, Beijing, China \\
$^2$Xi'an Jiaotong-Liverpool University, Suzhou, China
}

\begin{document}
\maketitle

\begingroup
\renewcommand\thefootnote{}
\footnotetext{This work was supported by the Suzhou Innovation and Entrepreneurship Leading Talents Programme - Innovation Leading Talent in Universities and Research Institutes with Grant No. ZXL2025310}
\endgroup

\begin{abstract}
LLM-based universal information extraction (UIE) methods often rely on additional information beyond the original training data, which increases training complexity yet often yields limited gains. To address this, we propose ProUIE, a Macro-to-Micro progressive learning approach that improves UIE without introducing any external information. ProUIE consists of three stages: (i) macro-level Complete Modeling (CM), which learns NER, RE, and EE along their intrinsic difficulty order on the full training data to build a unified extraction foundation, (ii) meso-level Streamlined Alignment (SA), which operates on sampled data with simplified target formats, streamlining and regularizing structured outputs to make them more concise and controllable, and (iii) micro-level Deep Exploration (DE), which applies GRPO with stepwise fine-grained rewards (SFR) over structural units to guide exploration and improve performance. Experiments on 36 public datasets show that ProUIE consistently improves unified extraction, outperforming strong instruction-tuned baselines on average for NER and RE while using a smaller backbone, and it further demonstrates clear gains in large-scale production-oriented information extraction.
\end{abstract}

\section{Introduction}
Universal information extraction (UIE) has emerged as a prominent direction in both academia and industry, aiming to unify diverse information extraction tasks under a unified framework. With the rapid progress of large language models (LLMs), numerous works on information extraction have adopted structured outputs, instruction tuning, and schema-driven formulations, moving beyond task-specific pipelines \citep{guideline, chatie, gptre, li-etal-2025-mpl} toward more general and flexible paradigms. However, many LLM-based UIE methods still rely on additional information beyond the original training data \citep{uie, schemaawareie, raguie, retrievalie, instructuie}, such as extra schema cues, external resources, or complex alignment and verification pipelines, thereby increasing training complexity while often yielding limited performance gains.

\begin{figure}[t]
  \centering
  \includegraphics[width=\linewidth]{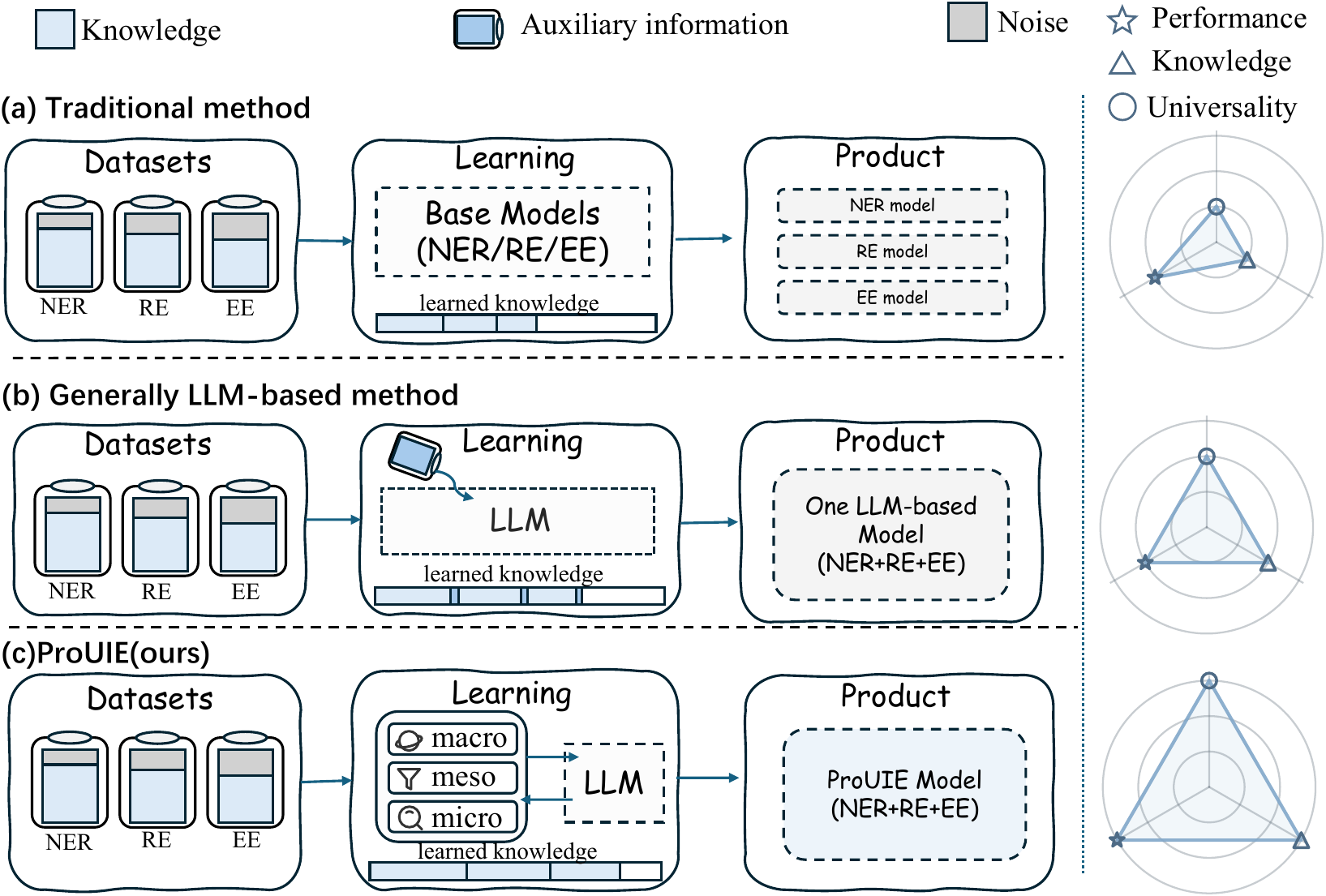}
  \caption{Comparison of UIE methods. (a) Traditional single-task methods train separate models for different extraction targets. (b) Generally LLM-based UIE unifies tasks but often introduces auxiliary signals beyond standard annotations. (c) ProUIE improves UIE without introducing any external information through a Macro-to-Micro progressive learning pipeline.}
  \label{fig:overview}
\end{figure}

As illustrated in Figure~\ref{fig:overview}, existing UIE methods can be broadly grouped into two paradigms: (a) Traditional single-task methods train and deploy separate models for different IE tasks, resulting in a straightforward pipeline but limited universality, and
(b) Generally LLM-based methods unify multiple targets into a single LLM via structured generation and instruction-based interfaces. Representative systems include schema-driven generation frameworks \citep{uie}, latent structure-aware generative modeling \citep{lasuie}, multi-task instruction tuning for unified extraction \citep{instructuie,yayiuie}, guideline-based supervision to enhance zero-shot extraction \citep{gollie}, and schema/knowledge coding with additional training stages \citep{knowcoder}. In addition, targeted distillation from large proprietary LLMs has been used to obtain broad NER ability under limited direct supervision \citep{universalner}. While these methods significantly advance UIE, many of them share a common design pattern: they introduce additional information beyond the original training data to stabilize learning and improve performance, thereby making the training pipeline heavier and less streamlined.

This naturally raises a key question: \textit{Is it possible to improve LLM-based UIE without introducing any external information?}
In this work, we answer this in the affirmative by rethinking the learning process itself: (i) how to progressively build a strong unified extraction foundation, (ii) how to make structured outputs concise and controllable with minimal overhead, and (iii) how to more fully exploit the model’s capacity through fine-grained optimization objectives. This motivates a training framework that strengthens UIE while keeping the supervision source unchanged and the pipeline simple.

To address this, we propose \textbf{ProUIE}, a macro-to-micro progressive learning framework that improves UIE without introducing any external information. ProUIE organizes training into three increasingly fine-grained stages: macro-level Complete Modeling (CM), meso-level Streamlined Alignment (SA), and micro-level Deep Exploration (DE). 

Specifically, the macro-level CM stage adopts supervised fine-tuning on the full training data using only standard annotations, serving as the foundation for unified extraction. Rather than relying on external information to constrain learning, CM aims to fully exploit the supervision contained in the original training data and establish a stable starting point for subsequent refinement. Built on CM, the meso-level SA stage further fine-tunes the model with a focus on the more complex RE/EE tasks, where structured generations tend to be verbose, inconsistent, or redundant. Concretely, SA randomly samples a subset of RE/EE training instances for focused refinement and trains the model to streamline outputs and remove redundancy, making the structured outputs more concise and controllable. Importantly, SA does not introduce any external resources or extra supervision; it improves controllability by tightening the output structure while staying within the standard-annotation regime. Finally, the micro-level DE stage targets structural errors that are difficult to resolve with conventional supervised learning. DE applies GRPO~\cite{shao2024deepseekmathpushinglimitsmathematical} together with our UIE-specific reward design, Stepwise Fine-grained Reward (SFR), which provides coarse-to-fine supervision over structural units to guide exploration. The key idea of SFR is to decompose structural correctness into progressively finer units and assign rewards across levels: for NER, SFR defines stepwise rewards at the entity-type and entity-value levels; for RE, at the relation-type, relation-instance, and entity-pair levels; and for EE, at the event-type, trigger, role, and argument levels. By shaping rewards from coarse to fine, DE can directly correct structural errors while still using only the original training data.

Our contributions are summarized as follows:
\begin{itemize}
\setlength\itemsep{0em}
  \item We propose \textbf{ProUIE}, a macro-to-micro progressive learning method that improves UIE without introducing any external information, and organizes training into three stages: \textbf{CM}, \textbf{SA}, and \textbf{DE}.
  \item We design \textbf{SFR}, a UIE-specific reward mechanism that provides stepwise fine-grained rewards over structural units, and integrate it with GRPO in the DE stage to guide exploration and further improve extraction quality.
  \item Extensive experiments on 36 public datasets demonstrate that ProUIE consistently outperforms strong baselines, and further delivers clear gains in large-scale production-oriented information extraction.

\end{itemize}

\section{Method}
\label{sec:method}

\subsection{Problem Definition}
Universal Information Extraction (UIE) aims to extract structured information from an input text $x$ under a unified generation interface, covering three tasks: named entity recognition (NER), relation extraction (RE), and event extraction (EE).
Given standard training data $\mathcal{D}=\{(x_i,y_i)\}_{i=1}^{N}$, where $y_i$ is the task-specific structured annotation, our goal is to learn a single LLM-based generator that produces structured outputs for all tasks \emph{without using any external auxiliary information}.

\begin{figure*}[t]
  \centering
  \includegraphics[width=\linewidth]{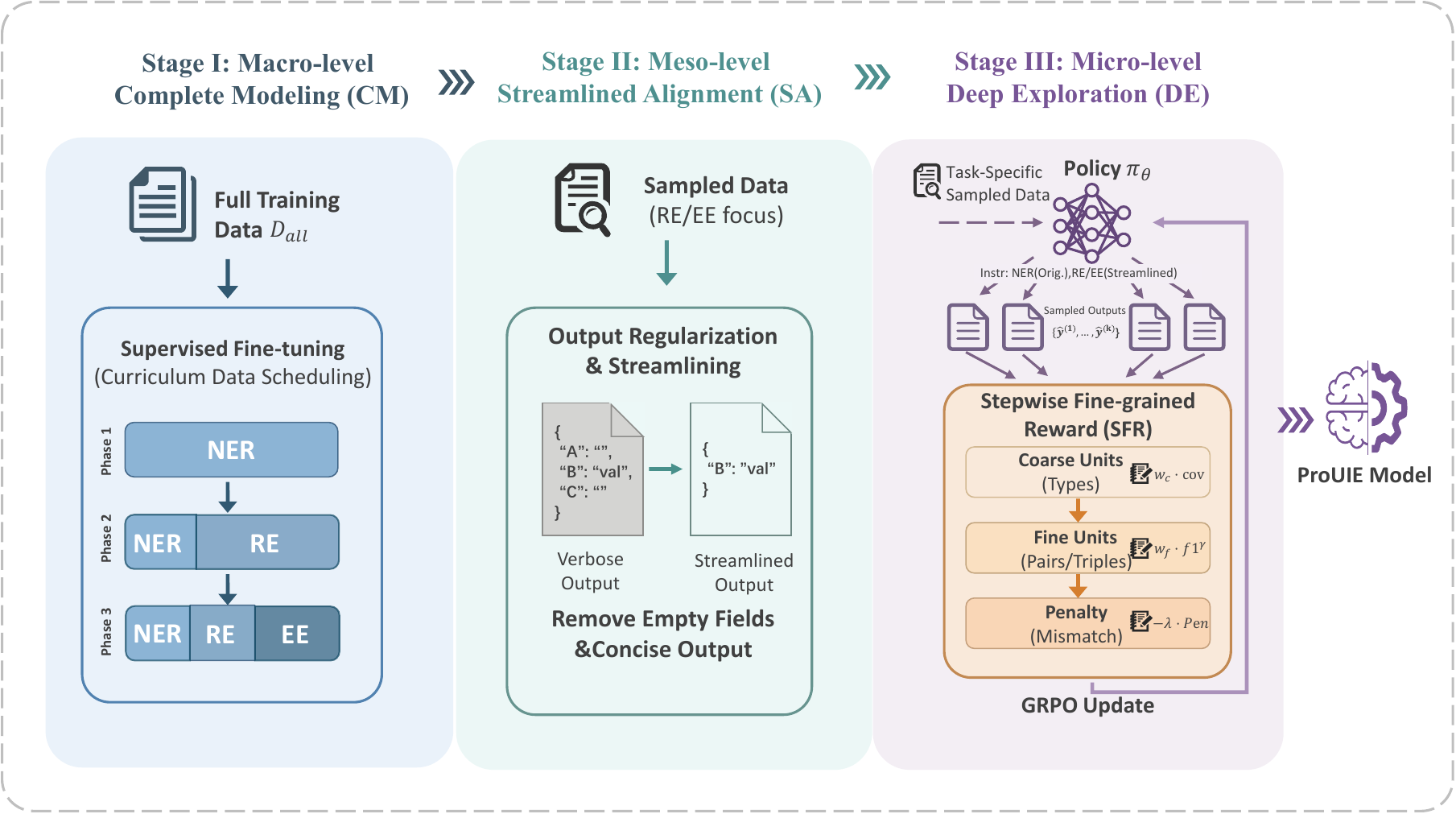}
  \caption{The framework of ProUIE.} 
  \label{fig:framework}
\end{figure*}

\subsection{Overview}
We propose \textbf{ProUIE}, a macro-to-micro progressive learning framework with three stages: \textbf{Complete Modeling (CM)}, \textbf{Streamlined Alignment (SA)}, and \textbf{Deep Exploration (DE)}.
CM builds a unified extraction foundation via task-ordered supervised training on full data.
SA improves controllability by fine-tuning on sampled data with streamlined targets.
DE further unlocks the model’s capacity and boosts final performance via GRPO with our \textbf{Stepwise Fine-grained Reward (SFR)}. Figure~\ref{fig:framework} illustrates the framework of ProUIE.

\subsection{Stage I: Complete Modeling (CM)}
The Complete Modeling (CM) stage establishes a unified extraction foundation for ProUIE.
At this macro level, we perform supervised fine-tuning on the full training data using only the original task annotations, without introducing any external signals or auxiliary supervision.

CM follows an intrinsic task-difficulty progression from NER to RE and EE, while maintaining unified generation behavior through mixed-task training.
Specifically, CM is organized into three phases.
In the first phase, the model is trained solely on NER data to acquire basic entity-level extraction capability.
In the second phase, NER and RE data are jointly used, with a sampling ratio of $2{:}8$, enabling the model to extend from entity recognition to relation modeling while retaining stable NER performance.
In the third phase, NER, RE, and EE data are trained together with a ratio of $3{:}3{:}4$, allowing the model to internalize increasingly complex extraction structures under a unified interface.

Rather than relying on external constraints, this phased design encourages the model to gradually absorb diverse extraction patterns while fully exploiting the supervision contained in the original training data.
As the first stage of ProUIE, CM provides a stable and general extraction starting point for subsequent stages that focus on output alignment and fine-grained optimization. We provide the CM prompt templates in Appendix~\ref{sec:appendix_prompts}.

\subsection{Stage II: Streamlined Alignment (SA)}
\label{sec:sa}
Built on CM, the meso-level \textbf{Streamlined Alignment (SA)} stage improves the \emph{controllability} of structured generation, especially for the more complex \textsc{RE}/\textsc{EE} tasks where outputs are often verbose.
SA does \emph{not} introduce any external resources.
Instead, it tightens the \emph{target format} while staying strictly within the original annotation regime.

Formally, SA fine-tunes the model on a randomly sampled subset $\mathcal{D}_{\textsc{sa}}\subset\mathcal{D}$ that primarily consists of \textsc{RE}/\textsc{EE} instances.
For each training example, we transform the original structured target into a \emph{streamlined} target by removing redundant fields and standardizing the representation.
For \textsc{RE}, we further enforce concise outputs by omitting empty relation fields.
For \textsc{EE}, we remove empty event-type entries, reducing invalid types and improving the conciseness and controllability of the outputs.
This encourages the model to produce concise, stable, and controllable outputs, which also provides a cleaner structural basis for the subsequent micro-level optimization in DE. We provide the SA prompt templates in Appendix~\ref{sec:appendix_prompts}.

\subsection{Stage III: Deep Exploration (DE)}
\label{subsec:de}

\subsubsection{GRPO with task-ordered phases}
After CM and SA, remaining errors are largely \emph{structural}: correct high-level types may be paired with wrong or missing slots, and redundant units may be generated.
DE addresses this by applying GRPO \cite{shao2024deepseekmathpushinglimitsmathematical} and rewarding sampled generations according to structural correctness.

To respect intrinsic task difficulty and stabilize optimization, DE proceeds in three GRPO phases:
\begin{equation}
\textsc{DE}_{\textsc{ner}}
\rightarrow
\textsc{DE}_{\textsc{re}}
\rightarrow
\textsc{DE}_{\textsc{ee}} .
\label{eq:de_order}
\end{equation}
In each phase, for every $(x,y)$ we sample $K$ candidates
$\{\hat{y}^{(k)}\}_{k=1}^{K}\sim\pi_{\theta}(\cdot\mid x)$,
compute rewards $\{R^{(k)}\}$, and update $\pi_\theta$ with GRPO using group-relative advantages. Optionally, the final reward can be clipped to $[0,1]$.

\subsubsection{Stepwise Fine-grained Reward (SFR)}
\label{subsec:sfr}
SFR evaluates a structured output from coarse units to fine units.
Coarse units encourage correct high-level structure (e.g., entity/relation/event types), while fine units focus on slot-level correctness (e.g., type--entity pairs, relation triples, event arguments).
A lightweight penalty discourages over/under-generation.

\paragraph{From outputs to structural units.}
We convert both gold output $y$ and prediction $\hat{y}$ into unit sets.
For each task we use one coarse set and one (or a few) fine sets:
\begin{itemize}
\setlength\itemsep{0.2em}
\item \textbf{NER:} coarse $T$ (entity types), fine $P$ (type--entity pairs).
\item \textbf{RE:} coarse $T$ (relation types), fine $R$ (type, head, tail triples),
and supportive fine sets $H$ (type, head) and $A$ (type, tail).
\item \textbf{EE:} coarse $E$ (event types) and $T$ (event type, trigger),
fine correctness measured by aligned role--argument pairs within matched trigger groups.
\end{itemize}

\paragraph{Reward form.}
Let $\mathrm{cov}(\cdot)$ denote coverage on a coarse set, $\mathrm{f1}(\cdot)$ denote micro-F1 on a fine set, and $\mathrm{pen}(\cdot)$ denote a mismatch penalty.
SFR follows a coarse-to-fine form:
\begin{equation}
R
=
w_c\,\mathrm{cov}
+
w_f\,(\mathrm{f1})^{\gamma}
-
\lambda\,\mathrm{pen},
\qquad \gamma>0.
\label{eq:sfr_form}
\end{equation}

\noindent\textbf{Definitions.}
Given gold set $A_g$ and predicted set $A_p$, we define:
\begin{equation}
\begin{aligned}
\mathrm{cov}(A_g,A_p)
&=
\frac{
|A_g\cap A_p|
}{
\max(1,|A_g|)
}.
\end{aligned}
\label{eq:cov_def}
\end{equation}

\begin{equation}
\begin{aligned}
\mathrm{f1}(A_g,A_p)
&=
\frac{
2|A_g\cap A_p|
}{
|A_g|+|A_p|
}.
\end{aligned}
\label{eq:f1_def}
\end{equation}

\begin{equation}
\Delta(A_g,A_p)=
\frac{|A_g\triangle A_p|}{\max(1,|A_g|)}.
\label{eq:delta_def}
\end{equation}

\begin{equation}
D_{\mathrm{jac}}(A_g,A_p)
=
1-
\frac{|A_g\cap A_p|}{\max(1,|A_g\cup A_p|)}.
\label{eq:jac_def}
\end{equation}

\paragraph{Instantiations for NER/RE/EE.}
We now specify $\mathrm{cov}$, $\mathrm{f1}$, and the penalty term for each task.

\textbf{NER.}
Coverage is computed on entity types, fine correctness is computed on type--entity pairs, penalties use normalized symmetric difference:
\begin{equation}
\begin{aligned}
R_{\textsc{sfr-ner}}
&=
w_t\,\mathrm{cov}(T_g,T_p)
+
w_p\,\big(\mathrm{f1}(P_g,P_p)\big)^{\gamma_{\textsc{ner}}}
\\
&\quad
-\lambda_t\,\Delta(T_g,T_p)
-\lambda_p\,\Delta(P_g,P_p).
\end{aligned}
\label{eq:ner_reward}
\end{equation}

\textbf{RE.}
Triple correctness is the main signal, head and tail F1 provides additional dense feedback.
We use Jaccard distance as a bounded mismatch penalty on types and triples:
\begin{equation}
\begin{aligned}
R_{\textsc{sfr-re}}
&=
w_t\,\mathrm{cov}(T_g,T_p)
+
w_h\,\mathrm{f1}(H_g,H_p)
\\
&\quad
+
w_a\,\mathrm{f1}(A_g,A_p)
+
w_r\,\big(\mathrm{f1}(R_g,R_p)\big)^{\gamma_{\textsc{re}}}
\\
&\quad
-\lambda_t\,D_{\mathrm{jac}}(T_g,T_p)
-\lambda_r\,D_{\mathrm{jac}}(R_g,R_p).
\end{aligned}
\label{eq:re_reward}
\end{equation}

\textbf{EE.}
We use event/trigger coverage as coarse signals.
For fine-grained correctness, we compute an aligned argument-level score $F_{\text{full}}$:
within each event type, each gold trigger group is matched to at most one predicted trigger group by maximum overlap of role--argument pairs, then we aggregate TP/FP/FN over role--argument pairs to obtain $F_{\text{full}}$.
We apply power stretching $\tilde{F}_{\text{full}}=(F_{\text{full}})^{\gamma_{\textsc{ee}}}$.
A piecewise penalty prioritizes coarse errors first (event mismatch, then trigger mismatch, then residual fine error):
\begin{equation}
\mathrm{pen}_{\text{ee}}=
\begin{cases}
\lambda_E\,\delta_E, & \delta_E>0,\\
\lambda_T\,\delta_T, & \delta_E=0\ \wedge\ \delta_T>0,\\
\lambda_F\,(1-F_{\text{full}}), & \delta_E=0\ \wedge\ \delta_T=0.
\end{cases}
\label{eq:ee_penalty}
\end{equation}
where $\delta_E=\Delta(E_g,E_p)$ and $\delta_T=\Delta(T_g,T_p)$.
The EE reward is:
\begin{equation}
\begin{aligned}
R_{\textsc{sfr-ee}}
&=
w_E\,\mathrm{cov}(E_g,E_p)
+
w_T\,\mathrm{cov}(T_g,T_p)
\\
&\quad
+
w_F\,\tilde{F}_{\text{full}}
-
\mathrm{pen}_{\text{ee}}.
\end{aligned}
\label{eq:ee_reward}
\end{equation}

\subsection{Training and Inference Procedure}
The learning procedure of ProUIE is presented in Algorithm~\ref{alg:prouie}, and its inference procedure is provided in Algorithm~\ref{alg:prouie_infer}.

\begin{algorithm}[t]
\caption{ProUIE Training Pipeline}
\label{alg:prouie}
\small
\KwIn{training data $\mathcal{D}$; base policy $\pi_{\theta}$; group size $K$.}
\KwOut{trained policy $\pi_{\theta}$.}

\BlankLine
\textbf{Stage I (CM):} Supervised fine-tuning on full $\mathcal{D}$ with original targets, following NER$\rightarrow$RE$\rightarrow$EE.\;

\textbf{Stage II (SA):} Sample $\mathcal{D}_{\textsc{sa}}\subset\mathcal{D}$ (mainly RE/EE); fine-tune with streamlined targets.\;

\textbf{Stage III (DE): Three GRPO phases}\;
\For(\tcp*[f]{Phase 1: NER}){each minibatch $\{(x,y)\}\sim\mathcal{D}_{\textsc{ner}}$}{
  \ForEach{$(x,y)$ in minibatch}{
    Sample $K$ outputs $\{\hat{y}^{(k)}\}_{k=1}^{K}\sim\pi_{\theta}(\cdot|x)$\;
    Compute rewards $\{R_{\textsc{sfr-ner}}^{(k)}\}$ via Eq.~\eqref{eq:ner_reward}\;
    Update $\pi_{\theta}$ with GRPO\;
  }
}
\For(\tcp*[f]{Phase 2: RE}){each minibatch $\{(x,y)\}\sim\mathcal{D}_{\textsc{re}}$}{
  \ForEach{$(x,y)$ in minibatch}{
    Sample $K$ outputs $\{\hat{y}^{(k)}\}_{k=1}^{K}\sim\pi_{\theta}(\cdot|x)$\;
    Compute rewards $\{R_{\textsc{sfr-re}}^{(k)}\}$ via Eq.~\eqref{eq:re_reward}\;
    Update $\pi_{\theta}$ with GRPO\;
  }
}
\For(\tcp*[f]{Phase 3: EE}){each minibatch $\{(x,y)\}\sim\mathcal{D}_{\textsc{ee}}$}{
  \ForEach{$(x,y)$ in minibatch}{
    Sample $K$ outputs $\{\hat{y}^{(k)}\}_{k=1}^{K}\sim\pi_{\theta}(\cdot|x)$\;
    Compute rewards $\{R_{\textsc{sfr-ee}}^{(k)}\}$ via Eq.~\eqref{eq:ee_reward}\;
    Update $\pi_{\theta}$ with GRPO\;
  }
}
\Return{$\pi_{\theta}$}\;
\end{algorithm}

\begin{algorithm}[t]
\caption{ProUIE Inference}
\label{alg:prouie_infer}
\small
\KwIn{input text $x$; task specification $\tau\in\{\textsc{NER},\textsc{RE},\textsc{EE}\}$.}
\KwOut{structured extraction output $\hat{y}$.}
Generate $\hat{y}\sim\pi_{\theta}(\cdot|x,\tau)$ using the trained model\;
\Return{$\hat{y}$}\;
\end{algorithm}

\section{Experiments}
\label{sec:exp}

We organize our experiments around the following research questions:

\begin{itemize}
\setlength\itemsep{0.2em}
  \item \textbf{RQ1:} How well does ProUIE perform compared with representative UIE baselines?
  \item \textbf{RQ2:} What are the contributions of each stage (CM/SA/DE) to the final performance?
  \item \textbf{RQ3:} Does SA improve generation efficiency by reducing the length of structured outputs?
  \item \textbf{RQ4:} Which reward granularity matters in DE, and how does the SFR affect learning?
  \item \textbf{RQ5:} How does ProUIE perform for large-scale real-world information extraction under strict schema-constrained requirements?
\end{itemize}

\begin{table*}[t]
\centering
\small
\setlength{\tabcolsep}{3.4pt}
\renewcommand{\arraystretch}{1.15}
\caption{NER results (F1, \%). Highest in each row is \textbf{bold}; second best is \underline{underlined}.}
\label{tab:ner_all}
\resizebox{\textwidth}{!}{
\begin{tabular}{lcccccccc}
\toprule
Dataset &
\makecell{BERT-base} &
\makecell{InstructUIE (11B)} &
\makecell{GollIE (7B)} &
\makecell{GollIE (13B)} &
\makecell{GollIE (34B)} &
\makecell{KnowCoder (7B)} &
\makecell{UniversalNER (7B)} &
\makecell{ProUIE (4B)} \\
\midrule
ACE2005 & 87.30 & 79.94 & 88.10 & \underline{89.40} & \textbf{89.60} & 86.10 & 86.69 & 86.15 \\
AnatEM & 85.82 & 88.52 & -- & -- & -- & 86.40 & \underline{88.65} & \textbf{89.09} \\
bc2gm & 80.90 & 80.69 & -- & -- & -- & 82.00 & \underline{82.42} & \textbf{83.87} \\
bc4chemd & 86.72 & 87.62 & -- & -- & -- & -- & \underline{89.21} & \textbf{91.97} \\
bc5cdr & 85.28 & \textbf{89.02} & 87.50 & 87.90 & 88.40 & \underline{89.30} & 89.34 & 88.93 \\
broad twitter & 58.61 & 80.27 & -- & -- & -- & 78.30 & \textbf{81.25} & \underline{79.27} \\
CoNLL2003 & 92.40 & 91.53 & 92.80 & 93.00 & 93.10 & \textbf{95.10} & \underline{93.30} & 92.41 \\
FabNER & 64.20 & 78.38 & -- & -- & -- & \textbf{82.90} & \underline{81.87} & 80.49 \\
FindVehicle & 87.13 & 87.56 & -- & -- & -- & \textbf{99.40} & \underline{98.30} & 98.00 \\
GENIA-Ent & 73.30 & 75.71 & -- & -- & -- & 76.70 & \underline{77.54} & \textbf{79.46} \\
HarveyNER & \textbf{82.26} & \underline{74.69} & -- & -- & -- & -- & 74.21 & 69.50 \\
multiNERD & 91.25 & 90.26 & -- & -- & -- & \textbf{96.10} & 93.73 & \underline{95.91} \\
ncbi-disease & 80.20 & 86.21 & 85.40 & 86.50 & 85.80 & 83.80 & \textbf{86.96} & \underline{86.95} \\
Ontonotes & \textbf{91.11} & 88.64 & 83.40 & 84.00 & 84.60 & 88.20 & \underline{89.91} & 88.33 \\
polyglot-NER & \textbf{75.65} & 53.31 & -- & -- & -- & -- & 65.67 & \underline{74.80} \\
tweetNER7 & 56.49 & \underline{65.95} & -- & -- & -- & -- & 65.77 & \textbf{66.44} \\
wikiann & 70.60 & 64.47 & -- & -- & -- & \textbf{87.00} & \underline{84.91} & 83.47 \\
wikineural & 82.78 & \underline{88.27} & -- & -- & -- & -- & \textbf{93.28} & 88.18 \\
Movie$^*$ & -- & \underline{63.00} & \underline{63.00} & 62.50 & 62.40 & 50.00 & 42.40 & \textbf{63.16} \\
Restaurant$^*$ & -- & 20.99 & 43.40 & 49.80 & \textbf{52.70} & \underline{48.20} & 31.70 & 52.41 \\
AI$^*$ & -- & 49.00 & 59.10 & 56.70 & \underline{61.60} & 60.30 & 53.50 & \textbf{62.36} \\
Literature$^*$ & -- & 47.21 & \underline{62.70} & 59.70 & 59.10 & 61.10 & 59.40 & \textbf{70.50} \\
Music$^*$ & -- & 53.16 & 56.20 & \underline{67.80} & 68.40 & 70.00 & 65.00 & \textbf{73.95} \\
Politics$^*$ & -- & 48.15 & 57.20 & 54.40 & 60.20 & \underline{72.20} & 60.80 & \textbf{78.67} \\
Science$^*$ & -- & 49.30 & 55.50 & 56.20 & 56.30 & 59.10 & \underline{61.10} & \textbf{61.84} \\
\midrule
Avg & -- & 71.27 & -- & -- & -- & -- & \underline{75.88} & \textbf{79.44} \\
\bottomrule
\end{tabular}}
\end{table*}

\subsection{Experimental Setup}
\label{sec:exp_setup}

\paragraph{Tasks and datasets.}
We evaluate three UIE tasks: NER, RE, and EE. Following prior UIE work, we report span-level micro-F1 for NER and RE, and report Trigger F1 / Argument F1 for EE. NER results are summarized in Table~\ref{tab:ner_all}, where datasets marked with $^\ast$ serve as out-of-domain (OOD) evaluations. RE and EE results are reported in Table~\ref{tab:re_results} and Tables~\ref{tab:ee_trigger}--\ref{tab:ee_argument}, respectively. Detailed dataset statistics are provided in Appendix ~\ref{sec:appendix_data_stats} (Table ~\ref{tab:data_stats_all}).\footnote{All datasets, models, and code used in this work are solely for academic research purposes and were not used for any commercial activities.}

\paragraph{Baselines.}
We compare ProUIE with representative UIE baselines, including traditional schema-driven UIE systems and instruction-tuned/generative UIE models (e.g., UIE~\citep{uie}, USM, InstructUIE~\citep{instructuie}, GollIE~\citep{gollie}, KnowCoder~\citep{knowcoder}, UniversalNER~\citep{universalner}). In our large-scale production-oriented extraction evaluation, we further compare against Qwen2.5 and Qwen3 \cite{yang2025qwen3} series models.

\paragraph{Evaluation metric.}
Following prior UIE work (e.g., \citet{instructuie}), we report span-level micro-F1 (\%) for NER/RE and Trigger/Argument F1 for EE. Unless otherwise stated, all results are from a single run with fixed hyperparameters.

\paragraph{Implementation details.}
We instantiate ProUIE as a 4B-parameter LLM-based extractor (\textsc{ProUIE (4B)}), built on the Qwen3-4B \cite{yang2025qwen3} backbone. More training details, prompts, and hyperparameters are provided in Appendix~\ref{sec:appendix_settings}.

\subsection{Comparison with Baselines (RQ1)}
\label{sec:exp_rq1}

\paragraph{NER.}
Table~\ref{tab:ner_all} reports NER results on diverse benchmarks, including OOD sets ($^\ast$). Overall, ProUIE achieves the best average score (\textbf{79.44}), surpassing \underline{UniversalNER} (+3.56 F1) and InstructUIE (+8.17 F1). It is particularly strong on biomedical datasets (e.g., \texttt{AnatEM}, \texttt{bc2gm}, \texttt{bc4chemd}, \texttt{GENIA-Ent}) while remaining competitive on general-domain benchmarks (e.g., \texttt{ACE2005}, \texttt{CoNLL2003}, \texttt{Ontonotes}). Moreover, ProUIE generalizes well out-of-domain, ranking best on 6/7 OOD datasets and second on \texttt{Restaurant}$^\ast$. Overall, these results support the effectiveness of our macro-to-micro progressive learning.

\begin{table}[t]
\centering
\small
\setlength{\tabcolsep}{5.2pt}
\renewcommand{\arraystretch}{1.15}
\caption{RE results (F1, \%). Highest in each row is bold; second best is underlined.}
\label{tab:re_results}
\resizebox{\columnwidth}{!}{
\begin{tabular}{lccccc}
\toprule
Dataset & UIE & USM & InstructUIE(11B) & KnowCoder(7B) & ProUIE(4B) \\
\midrule
ADE\_corpus & -- & -- & 82.31 & \textbf{84.30} & \textbf{84.30} \\
CoNLL2004  & 76.00 & \textbf{78.84} & \underline{78.48} & 73.30 & 73.17 \\
GIDS       & -- & -- & \textbf{81.98} & \underline{78.00} & 77.96 \\
kbp37      & -- & -- & 36.14 & \textbf{73.20} & \underline{69.45} \\
NYT        & -- & -- & 90.47 & \underline{93.70} & \textbf{96.41} \\
NYT11 HRL  & -- & -- & \textbf{56.06} & -- & \underline{52.68} \\
SciERC     & 36.53 & 37.36 & \underline{45.15} & 40.00 & \textbf{46.76} \\
Semeval RE & -- & -- & \textbf{73.23} & 66.30 & \underline{68.62} \\
\midrule
Avg        & -- & -- &\underline{67.98} & -- & \textbf{71.17}\\
\bottomrule
\end{tabular}
}
\end{table}

\paragraph{RE.}
Table~\ref{tab:re_results} reports RE results on multiple benchmarks. 
Overall, ProUIE attains the best average F1 among methods with reported Avg scores (\textbf{71.17}), outperforming \underline{InstructUIE} by +3.19 F1. 
Notably, ProUIE achieves these results with a smaller backbone (ProUIE 4B vs. InstructUIE 11B / KnowCoder 7B) and without introducing any additional auxiliary information beyond the original training data. 
ProUIE is particularly strong on benchmarks with richer schemas and diverse relation patterns, such as \texttt{NYT} (96.41) and \texttt{SciERC} (46.76). 
Meanwhile, other baselines remain competitive on several datasets (e.g., \texttt{Semeval RE}), suggesting that performance varies with dataset characteristics.

\paragraph{EE.}
Table~\ref{tab:ee_trigger} and Table~\ref{tab:ee_argument} report event extraction results on trigger identification and argument extraction, respectively. Overall, ProUIE shows mixed performance on EE. On \texttt{PHEE}, ProUIE achieves the best argument F1 (71.86), indicating that the proposed training pipeline can benefit complex argument extraction in certain domains. However, on \texttt{ACE2005} and \texttt{CASIE}, ProUIE underperforms the strongest baseline for both triggers and arguments, suggesting that EE remains challenging under our current unified generation setting and reward configuration. We leave improving EE stability and closing the gap on general-domain EE benchmarks as an important direction for future work.

\begin{table}[h]
\centering
\small
\setlength{\tabcolsep}{3.8pt}
\renewcommand{\arraystretch}{1.05}
\caption{EE Trigger Identification results (F1, \%). }
\label{tab:ee_trigger}
\resizebox{\columnwidth}{!}{
\begin{tabular}{lccccc}
\toprule
Dataset & BERT-base & UIE & USM & InstructUIE & ProUIE \\
\midrule
ACE2005 & 72.50 & \underline{73.36} & 72.41 & \textbf{77.13} & 70.65 \\
CASIE   & 68.98 & \underline{69.33} & \textbf{71.73} & 67.80 & 54.17 \\
PHEE    & --    & --    & --    & \textbf{70.14} & \underline{67.87} \\
\bottomrule
\end{tabular}}
\end{table}

\begin{table}[h]
\centering
\small
\setlength{\tabcolsep}{3.8pt}
\renewcommand{\arraystretch}{1.05}
\caption{EE Argument Extraction results (F1, \%). }
\label{tab:ee_argument}
\resizebox{\columnwidth}{!}{
\begin{tabular}{lccccc}
\toprule
Dataset & BERT-base & UIE & USM & InstructUIE & ProUIE \\
\midrule
ACE2005 & \underline{59.90} & 54.79 & 55.83 & \textbf{72.94} & 52.58 \\
CASIE   & 60.37 & 61.30 & \underline{63.26} & \textbf{63.53} & 60.75 \\
PHEE    & --    & --    & --    & \underline{62.91} & \textbf{71.86} \\
\bottomrule
\end{tabular}}
\end{table}

\subsection{Stage-wise Ablation (RQ2)}
\label{sec:exp_rq2}

To quantify the contribution of each stage, we perform stage-wise ablations:
\begin{itemize}
\setlength\itemsep{0.2em}
  \item \textbf{w/o SA:} remove Stage II (SA), i.e., train CM$\rightarrow$DE directly.
  \item \textbf{w/o DE:} remove Stage III (DE), i.e., stop after CM$\rightarrow$SA.
  \item \textbf{CM only:} use only Stage I (CM) training.
\end{itemize}

\paragraph{Results.}
Table~\ref{tab:ablation_stage} shows that each stage contributes to the final performance. 
Removing DE results in the largest average degradation across the three tasks, with clear drops on NER and notable declines on RE and EE, highlighting the importance of our SFR-based GRPO refinement. 
Removing SA also consistently hurts performance, particularly on RE and EE, suggesting that streamlining stabilizes structured generation and provides a better starting point for DE.

\begin{table}[H]
\centering
\small
\setlength{\tabcolsep}{5.0pt}
\renewcommand{\arraystretch}{1.12}
\caption{Stage-wise ablation study of ProUIE.}
\label{tab:ablation_stage}

\newcommand{\hangcell}[1]{\hangindent=1.2em\hangafter=1 #1}

\begin{tabular}{@{} >{\raggedright\arraybackslash}p{0.50\columnwidth} c c c @{}}
\toprule
Method & NER & RE & EE \\
\midrule
\hangcell{ProUIE (CM+SA+DE)} & \textbf{79.44} & \textbf{71.17} & \textbf{61.73} \\
\hangcell{w/o SA (CM+DE)}            & 78.19 & 68.68  & 57.92 \\
\hangcell{w/o DE (CM+SA)}            & 76.22 & 68.26 & 58.50 \\
\hangcell{CM only}                   & 76.27 & 70.56  & 58.07 \\
\bottomrule
\end{tabular}
\end{table}

\subsection{Analysis of Streamlined Alignment (SA)}
\label{sec:exp_rq3}

We examine whether SA improves generation efficiency by shortening structured outputs for \textsc{RE} and \textsc{EE}.
For this analysis, \textsc{RE} uses \texttt{SciERC} and \texttt{SemEval RE}, and \textsc{EE} uses \texttt{ACE2005} and \texttt{CASIE}.
We report token-length distributions using percentiles (P50/P70/P99), where each bucket summarizes the min/mean/max token length among examples below the corresponding threshold.

\paragraph{Results.}
Table~\ref{tab:sa_tokens_dist} shows that SA substantially compresses outputs for both tasks across the entire distribution.
For \textsc{RE}, the mean length drops from 70/83/114 tokens (P50/P70/P99) to 13/18/74, and the P99 maximum decreases from 161 to 126.
For \textsc{EE}, the mean length decreases from 81/105/171 to 28/37/90, and the P99 maximum is nearly halved, from 199 to 99.
These consistent reductions indicate that SA effectively removes redundancy and stabilizes formatting, leading to lower decoding cost in both training (e.g., sampling in DE) and inference.

\begin{figure*}[t]
  \centering
  \includegraphics[width=\linewidth]{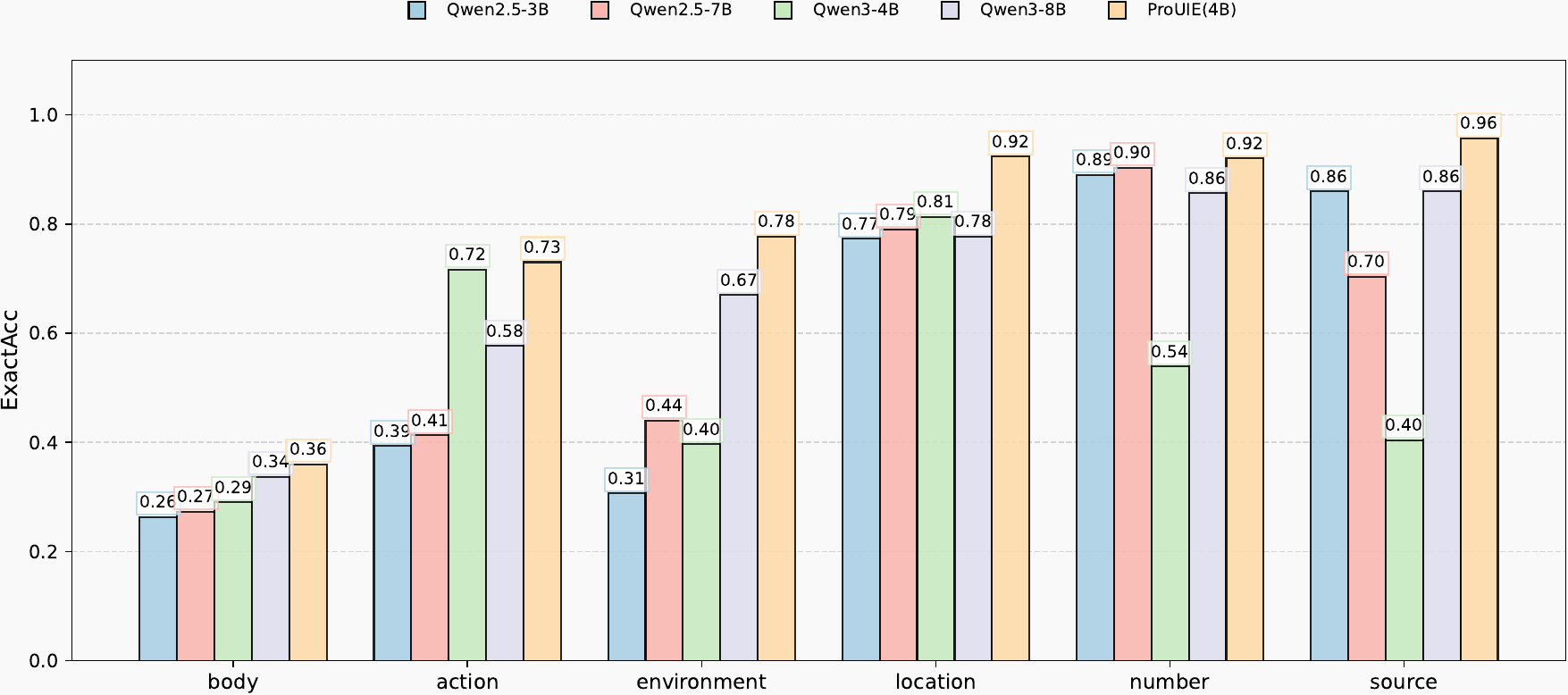}
  \caption{Production-scale mobile query understanding: slot-level exact-match accuracy on six representative fields.}
  \label{fig:nlu}
\end{figure*}

\begin{table}[t]
\centering
\small
\setlength{\tabcolsep}{3.2pt}      
\renewcommand{\arraystretch}{1.08}
\caption{Token-length statistics of generated outputs.}
\label{tab:sa_tokens_dist}
\resizebox{\columnwidth}{!}{
\begin{tabular}{llccc|ccc|ccc}
\toprule
\multirow{2}{*}{Task} & \multirow{2}{*}{Setting} &
\multicolumn{3}{c|}{P50} & \multicolumn{3}{c|}{P70} & \multicolumn{3}{c}{P99} \\
\cmidrule(lr){3-5}\cmidrule(lr){6-8}\cmidrule(lr){9-11}
& & min & mean & max & min & mean & max & min & mean & max \\
\midrule
\multirow{2}{*}{RE}
& CM    & 58 & 70 & 75 & 76 & 83 & 85 & 100 & 114 & 161 \\
& CM+SA & 10 & 13 & 20 & 14 & 18 & 35 & 39 & 74 & 126 \\
\midrule
\multirow{2}{*}{EE}
& CM    & 45 & 81 & 150 & 62 & 105 & 170 & 115 & 171 & 199 \\
& CM+SA & 16 & 28 & 34 & 18 & 37 & 42 & 48 & 90 & 99 \\
\bottomrule
\end{tabular}}
\end{table}

\subsection{Effect of SFR Reward Granularity in DE (RQ4)}
\label{sec:exp_rq4}

DE applies GRPO with our stepwise fine-grained reward (\textbf{SFR}). We study the effect of reward granularity by ablating SFR into the following variants:
\begin{itemize}
\setlength\itemsep{0.2em}
  \item \textbf{Coarse-only:} only coarse-grained rewards.
  \item \textbf{Fine-only:} only fine-grained rewards.
  \item \textbf{Coarse-to-fine (SFR):} full SFR combining coarse and fine rewards.
\end{itemize}

\paragraph{Results.}
Table~\ref{tab:reward_ablation} shows that \textbf{coarse-only} rewards underperform, as they mainly encourage high-level correctness but provide limited signal for fixing slot-level errors. 
\textbf{Fine-only} rewards improve over coarse-only by directly optimizing fine units, yet remain sensitive to early mismatches (e.g., incorrect types or grouping), which can hinder stable refinement. 
Overall, our \textbf{coarse-to-fine SFR} achieves the best performance across NER, RE, and EE, indicating that combining coarse guidance with fine-grained matching yields more effective and robust GRPO optimization.

\begin{table}[t]
\centering
\small
\setlength{\tabcolsep}{5.0pt}
\renewcommand{\arraystretch}{1.12}
\caption{Coarse-to-fine reward ablation in DE.}
\label{tab:reward_ablation}
\begin{tabular}{lccc}
\toprule
Reward Granularity & NER (Avg) & RE (Avg) & EE (Avg) \\
\midrule
Coarse-only           & 77.43 & 65.88 & 57.14 \\
Fine-only    & 77.86 & 67.44& 57.84 \\
Coarse-to-fine SFR       & \textbf{78.25} & \textbf{67.69} & \textbf{58.40} \\
\bottomrule
\end{tabular}
\end{table}

\subsection{Case Study}
\label{sec:exp_case}

We present qualitative examples to illustrate how each stage improves structured prediction. Appendix \ref{sec:appendix_case} shows representative samples where (i) CM captures the general extraction pattern but produces redundant or inconsistent structures, (ii) SA makes the structure concise and controllable, and (iii) DE further corrects hard structural errors.

\subsection{Production-oriented Evaluation (RQ5)}
\label{sec:exp_business}

We evaluate ProUIE on a large-scale production-oriented query understanding task with strict JSON-constrained slot extraction. Due to confidentiality, we report \textit{ExactAcc} on six representative slots (\texttt{body}, \texttt{action}, \texttt{environment}, \texttt{location}, \texttt{number}, \texttt{source}). As shown in Figure~\ref{fig:nlu}, ProUIE consistently outperforms the baselines across all reported slots, demonstrating better alignment to user-facing schema constraints.
Details are in Appendix~\ref{sec:appendix_business}.

\section{Related Work}
\label{sec:related_work}
\paragraph{Task-specific information extraction models.}

Conventional IE research typically develops dedicated models for individual tasks such as Named Entity Recognition (NER), Relation Extraction (RE), and Event Extraction (EE), optimizing task-specific architectures and objectives~\cite{ner, seqner, re, ee}. These models often achieve strong in-domain performance, but they require separate training and deployment for different targets, and their interfaces are not unified across tasks.

\paragraph{LLM-based universal information extraction.}
Universal information extraction (UIE) aims to unify heterogeneous IE tasks under a single modeling and inference framework, often by casting extraction as structured generation. Representative work formulates UIE as schema-driven generation \citep{uie} and extends it with instruction tuning to unify multiple extraction targets using LLMs \citep{instructuie, yayiuie, chatie}. More recent LLM-based UIE approaches introduce richer guidance signals or additional training stages to improve generalization and controllability, such as guideline-driven supervision \citep{gollie} and schema/knowledge coding \citep{shemadriven, knowcoder}. 
In contrast, ProUIE improves UIE without introducing external auxiliary information beyond the original training data, and instead strengthens unified extraction via macro-to-micro progressive learning method.

\section{Conclusion}
In this paper, we propose ProUIE, a macro-to-micro progressive learning framework for LLM-based universal information extraction that improves UIE without introducing external auxiliary information beyond the original training data. ProUIE consists of three stages: Complete Modeling (CM), Streamlined Alignment (SA), and Deep Exploration (DE). CM builds a unified extraction foundation with task-ordered supervised training, SA streamlines and stabilizes structured outputs, and DE applies GRPO with our coarse-to-fine Stepwise Fine-grained Reward (SFR) to refine structural correctness. Experiments and analyses demonstrate the effectiveness of ProUIE, and future work will focus on further improving event extraction and reward design.

\section*{Limitations}
ProUIE shows less consistent improvements on event extraction (EE) across benchmarks, especially under a relatively small backbone setting (4B). This indicates that learning stable event structures under a unified generation interface remains challenging, and we expect further progress to mainly come from stronger event-specific modeling and more appropriate reward design.

In addition, ProUIE adopts a multi-stage training pipeline (CM$\rightarrow$SA$\rightarrow$DE), where the DE stage relies on GRPO with sampled generations. Although this design does not introduce external supervision beyond the original training data, it incurs additional computational cost compared to pure supervised fine-tuning, which may limit applicability in settings with constrained training budgets.


\bibliography{ProUIE_ref}

\clearpage
\appendix
\section{Appendix}
\label{sec:appendix}

\subsection{Dataset Statistics}
\label{sec:appendix_data_stats}

We summarize detailed statistics of all datasets used in our experiments, including the number of labels and the train/validation/test split sizes for each task, in Table~\ref{tab:data_stats_all}. 
For EE, the number of event types is reported, with argument roles in parentheses.

\begin{table}[!htbp]
\centering
\small
\setlength{\tabcolsep}{3.0pt}      
\renewcommand{\arraystretch}{1.05}
\caption{Detailed dataset statistics. $\lvert\mathcal{Y}\rvert$ denotes the number of labels.}
\label{tab:data_stats_all}

\resizebox{\columnwidth}{!}{ 
\begin{tabular}{@{} l p{0.34\columnwidth} r r r r @{}}
\toprule
Task & Dataset & $|\mathcal{Y}|$ & \#Train & \#Val & \#Test \\
\midrule

\multirow{24}{*}{\textsc{NER}}
& ACE2005               & 7   & 15,000   & 971    & 1,060  \\
& broad\_twitter\_corpus & 3   & 5,334    & 2,000  & 2,001  \\
& CoNLL2003             & 4   & 15,000   & 3,250  & 3,453  \\
& multiNERD             & 16  & 134,144  & 10,000 & 10,000 \\
& Ontonotes             & 18  & 30,000   & 8,528  & 8,262  \\
& polyglot-NER          & 3   & 393,982  & 10,000 & 10,000 \\
& tweetNER7             & 7   & 15,000   & 886    & 576    \\
& wikiann               & 3   & 20,000   & 10,000 & 10,000 \\
& wikineural            & 3   & 92,720   & 11,590 & 11,597 \\
& AnatEM                & 1   & 5,861    & 2,118  & 3,830  \\
& bc2gm                 & 1   & 12,500   & 2,500  & 5,000  \\
& bc4chemd              & 1   & 30,682   & 30,639 & 26,364 \\
& bc5cdr                & 2   & 4,560    & 4,581  & 4,797  \\
& CrossNER\_AI          & 14  & --       & 350    & 431    \\
& CrossNER\_literature  & 12  & --       & 400    & 416    \\
& CrossNER\_music       & 13  & --       & 380    & 465    \\
& CrossNER\_politics    & 9   & --       & 540    & 650    \\
& CrossNER\_science     & 17  & --       & 450    & 543    \\
& FabNER                & 12  & 9,435    & 2,182  & 2,064  \\
& FindVehicle           & 21  & 21,565   & 20,777 & 20,777 \\
& GENIA                 & 5   & 15,023   & 1,669  & 1,854  \\
& HarveyNER             & 4   & 3,967    & 1,301  & 1,303  \\
& MIT\_Movie            & 12  & --       & 2,442  & 2,442  \\
& MIT\_Restaurant       & 8   & --       & 1,520  & 1,520  \\
& ncbi-disease          & 1   & 5,432    & 923    & 940    \\

\midrule
\multirow{8}{*}{\textsc{RE}}
& ADE\_corpus  & 1   & 3,417   & 427   & 428   \\
& CoNLL2004    & 5   & 922     & 231   & 288   \\
& GIDS         & 4   & 8,526   & 1,417 & 4,307 \\
& kbp37        & 18  & 15,917  & 1,724 & 3,405 \\
& NYT          & 24  & 56,190  & 5,000 & 5,000 \\
& NYT11 HRL    & 12  & 62,648  & 149   & 369   \\
& SciERC       & 7   & 1,366   & 187   & 397   \\
& semeval RE   & 10  & 6,507   & 1,493 & 2,717 \\

\midrule
\multirow{3}{*}{\textsc{EE}}
& ACE2005 & 33(22) & 3,342 & 327 & 293   \\
& CASIE   & 5(26)  & 3,751 & 788 & 1,500 \\
& PHEE    & 2(16)  & 2,898 & 961 & 968   \\

\bottomrule
\end{tabular}}
\end{table}

\subsection{Prompt Templates}
\label{sec:appendix_prompts}

For reproducibility, we provide the prompt templates used in Stage I (CM) and Stage II (SA), referenced in the main text as Figure~\ref{fig:prompt-ner-instruction}--Figure~\ref{fig:prompt-ee-instruction} and Figure~\ref{fig:prompt-re-concise}--Figure~\ref{fig:prompt-ee-concise}.

\begingroup
\setlength{\textfloatsep}{4pt}
\setlength{\floatsep}{3pt}
\setlength{\intextsep}{3pt}
\setlength{\dbltextfloatsep}{4pt}
\setlength{\dblfloatsep}{3pt}

\renewcommand{\topfraction}{0.99}
\renewcommand{\bottomfraction}{0.99}
\renewcommand{\textfraction}{0.01}
\renewcommand{\floatpagefraction}{0.90}
\renewcommand{\dblfloatpagefraction}{0.90}

\setcounter{topnumber}{50}
\setcounter{bottomnumber}{50}
\setcounter{totalnumber}{50}
\setcounter{dbltopnumber}{50}

\lstset{
  basicstyle=\ttfamily\footnotesize\linespread{1.08}\selectfont,
  breaklines=true,
  breakatwhitespace=true,
  columns=fullflexible,
  keepspaces=true,
  frame=none,
  breakindent=0.30em,
  breakautoindent=false,
  postbreak=\mbox{},
  aboveskip=0pt,
  belowskip=0pt
}

\begin{figure}[!p]
\centering
\begin{tcolorbox}[
  enhanced,
  colback=white,
  colframe=black!55,
  boxrule=0.55pt,
  arc=1.2pt,
  left=1.2mm,right=1.2mm,   
  top=0.9mm,bottom=0.9mm,   
  boxsep=0.4mm             
]

\begin{tcolorbox}[
  enhanced,
  colback=white,
  colframe=black!55,
  boxrule=0.45pt,
  arc=1.0pt,
  left=1.4mm,right=1.4mm,
  top=0.7mm,bottom=0.7mm,
  boxsep=0.6mm,
  title={Instruction},
  fonttitle=\bfseries\footnotesize,
  colbacktitle=black!12,
  coltitle=black,
  titlerule=0.45pt
]
\begin{lstlisting}
You are an information extraction assistant. Strictly extract {len(slots)} slots ({', '.join(slots)}) from the user input. Slot values must be exact substrings of the input. If a slot has multiple values, join them with ' | ' in the order of first appearance and remove duplicates. The user input is:
\end{lstlisting}
\end{tcolorbox}

\vspace{0.25mm} 

\begin{tcolorbox}[
  enhanced,
  colback=white,
  colframe=black!55,
  boxrule=0.45pt,
  arc=1.0pt,
  left=1.4mm,right=1.4mm,
  top=0.7mm,bottom=0.7mm,
  boxsep=0.6mm,
  title={Input Example},
  fonttitle=\bfseries\footnotesize,
  colbacktitle=black!12,
  coltitle=black,
  titlerule=0.45pt
]
\vspace{1.2mm} 
\begin{lstlisting}
Best wishes to Kevin , Therese & their family as they embark on the next stage of their lives . JG
\end{lstlisting}
\end{tcolorbox}

\vspace{0.25mm}

\begin{tcolorbox}[
  enhanced,
  colback=white,
  colframe=black!55,
  boxrule=0.45pt,
  arc=1.0pt,
  left=1.4mm,right=1.4mm,
  top=0.7mm,bottom=0.7mm,
  boxsep=0.6mm,
  title={Output Example},
  fonttitle=\bfseries\footnotesize,
  colbacktitle=black!12,
  coltitle=black,
  titlerule=0.45pt
]
\vspace{0.25mm}
\begin{lstlisting}
{
  "location": "", 
  "person": "Kevin | Therese", 
  "organization": ""
}
\end{lstlisting}
\end{tcolorbox}

\end{tcolorbox}
\vspace{-5mm}
\caption{Prompt template for NER fine-tuning in CM stage.}

\label{fig:prompt-ner-instruction}
\end{figure}

\lstset{
  basicstyle=\ttfamily\footnotesize\linespread{1.08}\selectfont,
  breaklines=true,
  breakatwhitespace=true,
  columns=fullflexible,
  keepspaces=true,
  frame=none,
  breakindent=0.30em,
  breakautoindent=false,
  postbreak=\mbox{},
  aboveskip=0pt,
  belowskip=0pt
}

\begin{figure}[!p]
\centering
\begin{tcolorbox}[
  enhanced,
  colback=white,
  colframe=black!55,
  boxrule=0.55pt,
  arc=1.2pt,
  left=1.2mm,right=1.2mm,
  top=0.9mm,bottom=0.9mm,
  boxsep=0.4mm
]

\begin{tcolorbox}[
  enhanced,
  colback=white,
  colframe=black!55,
  boxrule=0.45pt,
  arc=1.0pt,
  left=1.4mm,right=1.4mm,
  top=0.7mm,bottom=4.5mm,
  boxsep=0.6mm,
  title={Instruction},
  fonttitle=\bfseries\footnotesize,
  colbacktitle=black!12,
  coltitle=black,
  titlerule=0.45pt
]
\begin{lstlisting}
You are an information extraction assistant. Strictly extract relation pairs for the following relation types ({", ".join(relation_types)}) from the user input. Values must be exact substrings of the input. If a relation has multiple pairs, join them with ' | ' in the order of first appearance, and each pair must be formatted as 'word1, word2'. The user input is:

\end{lstlisting}
\end{tcolorbox}

\vspace{0.25mm}

\begin{tcolorbox}[
  enhanced,
  colback=white,
  colframe=black!55,
  boxrule=0.45pt,
  arc=1.0pt,
  left=1.4mm,right=1.4mm,
  top=0.7mm,bottom=0.7mm,
  boxsep=0.6mm,
  title={Input Example},
  fonttitle=\bfseries\footnotesize,
  colbacktitle=black!12,
  coltitle=black,
  titlerule=0.45pt
]
\vspace{1.2mm}
\begin{lstlisting}
Instead of representing scene/object by a collection of isolated 3D features -LRB- usually points -RRB- , our algorithm uses a surface controlled by a small set of parameters .
\end{lstlisting}
\end{tcolorbox}

\vspace{0.25mm}

\begin{tcolorbox}[
  enhanced,
  colback=white,
  colframe=black!55,
  boxrule=0.45pt,
  arc=1.0pt,
  left=1.4mm,right=1.4mm,
  top=0.7mm,bottom=0.7mm,
  boxsep=0.6mm,
  title={Output Example},
  fonttitle=\bfseries\footnotesize,
  colbacktitle=black!12,
  coltitle=black,
  titlerule=0.45pt
]
\vspace{0.25mm}
\begin{lstlisting}
{
  "conjunction": "",
  "feature of": "",
  "hyponym of": "",
  "used for": "surface, algorithm",
  "part of": "",
  "compare": "",
  "evaluate for": ""
}

\end{lstlisting}

\end{tcolorbox}

\end{tcolorbox}
\vspace{-5mm}
\caption{Prompt template for RE fine-tuning in CM stage.}
\label{fig:prompt-re-instruction}
\end{figure}

\lstset{
  basicstyle=\ttfamily\footnotesize\linespread{1.08}\selectfont,
  breaklines=true,
  breakatwhitespace=true,
  columns=fullflexible,
  keepspaces=true,
  frame=none,
  breakindent=0.30em,
  breakautoindent=false,
  postbreak=\mbox{},
  aboveskip=0pt,
  belowskip=0pt
}

\begin{figure*}[!p]
\centering
\begin{tcolorbox}[
  enhanced,
  colback=white,
  colframe=black!55,
  boxrule=0.55pt,
  arc=1.2pt,
  left=1.2mm,right=1.2mm,
  top=0.9mm,bottom=0.9mm,
  boxsep=0.4mm
]

\begin{tcolorbox}[
  enhanced,
  colback=white,
  colframe=black!55,
  boxrule=0.45pt,
  arc=1.0pt,
  left=1.4mm,right=1.4mm,
  top=0.7mm,bottom=0.7mm,
  boxsep=0.6mm,
  title={Instruction},
  fonttitle=\bfseries\footnotesize,
  colbacktitle=black!12,
  coltitle=black,
  titlerule=0.45pt
]
\begin{lstlisting}
You are an information extraction assistant. Strictly extract events for the following event types from the user input and reply with a single JSON object only. The keys MUST be exactly these event types: [<EVENT_TYPES>]. For each event type, group arguments by trigger. Format each group as `TRIGGER: ROLE: argument; ROLE: argument`. Valid roles include: [<ROLE_TYPES>]. The user input is:
\end{lstlisting}
\end{tcolorbox}

\vspace{0.25mm}

\begin{tcolorbox}[
  enhanced,
  colback=white,
  colframe=black!55,
  boxrule=0.45pt,
  arc=1.0pt,
  left=1.4mm,right=1.4mm,
  top=0.7mm,bottom=0.7mm,
  boxsep=0.6mm,
  title={Input Example},
  fonttitle=\bfseries\footnotesize,
  colbacktitle=black!12,
  coltitle=black,
  titlerule=0.45pt
]
\vspace{1.2mm}
\begin{lstlisting}
After 5 days of treatment with IL-2, the patient developed a hemorrhagic lesion that progressed to toxic epidermal necrolysis, as well as grade 4 pancytopenia.
\end{lstlisting}
\end{tcolorbox}

\vspace{0.25mm}

\begin{tcolorbox}[
  enhanced,
  colback=white,
  colframe=black!55,
  boxrule=0.45pt,
  arc=1.0pt,
  left=1.4mm,right=1.4mm,
  top=0.7mm,bottom=0.7mm,
  boxsep=0.6mm,
  title={Output Example},
  fonttitle=\bfseries\footnotesize,
  colbacktitle=black!12,
  coltitle=black,
  titlerule=0.45pt
]
\vspace{0.25mm}
\begin{lstlisting}
{
"potential therapeutic event": "",
"adverse event": "developed: Subject: patient; Effect: a hemorrhagic lesion that progressed to toxic epidermal necrolysis, as well as grade 4 pancytopenia; Treatment: 5 days of treatment with IL-2; Treatment.Drug: IL-2; Treatment.Time_elapsed: After 5 days"
}
\end{lstlisting}
\end{tcolorbox}

\end{tcolorbox}
\vspace{-5mm}
\caption{Prompt template for EE fine-tuning in CM stage.}
\label{fig:prompt-ee-instruction}
\end{figure*}

\lstset{
  basicstyle=\ttfamily\footnotesize\linespread{1.08}\selectfont,
  breaklines=true,
  breakatwhitespace=true,
  columns=fullflexible,
  keepspaces=true,
  frame=none,
  breakindent=0.30em,
  breakautoindent=false,
  postbreak=\mbox{},
  aboveskip=0pt,
  belowskip=0pt
}

\begin{figure*}[!p]
\centering
\begin{tcolorbox}[
  enhanced,
  colback=white,
  colframe=black!55,
  boxrule=0.55pt,
  arc=1.2pt,
  left=1.2mm,right=1.2mm,
  top=0.9mm,bottom=0.9mm,
  boxsep=0.4mm
]

\begin{tcolorbox}[
  enhanced,
  colback=white,
  colframe=black!55,
  boxrule=0.45pt,
  arc=1.0pt,
  left=1.4mm,right=1.4mm,
  top=0.7mm,bottom=0.7mm,
  boxsep=0.6mm,
  title={Instruction},
  fonttitle=\bfseries\footnotesize,
  colbacktitle=black!12,
  coltitle=black,
  titlerule=0.45pt
]
\begin{lstlisting}
You are an information extraction assistant...
Please use concise output with no empty fields. The user input is:
\end{lstlisting}

\end{tcolorbox}

\vspace{0.25mm}

\begin{tcolorbox}[
  enhanced,
  colback=white,
  colframe=black!55,
  boxrule=0.45pt,
  arc=1.0pt,
  left=1.4mm,right=1.4mm,
  top=0.7mm,bottom=0.7mm,
  boxsep=0.6mm,
  title={Input Example},
  fonttitle=\bfseries\footnotesize,
  colbacktitle=black!12,
  coltitle=black,
  titlerule=0.45pt
]
\vspace{1.2mm}
\begin{lstlisting}
..., our algorithm uses a surface controlled by a small set of parameters .
\end{lstlisting}
\end{tcolorbox}

\vspace{0.25mm}

\begin{tcolorbox}[
  enhanced,
  colback=white,
  colframe=black!55,
  boxrule=0.45pt,
  arc=1.0pt,
  left=1.4mm,right=1.4mm,
  top=0.7mm,bottom=0.7mm,
  boxsep=0.6mm,
  title={Output Example},
  fonttitle=\bfseries\footnotesize,
  colbacktitle=black!12,
  coltitle=black,
  titlerule=0.45pt
]
\vspace{0.25mm}
\begin{lstlisting}
{"used for": "surface, algorithm"}
\end{lstlisting}
\end{tcolorbox}

\end{tcolorbox}
\vspace{-5mm}
\caption{Concise RE output for SA stage.}
\label{fig:prompt-re-concise}
\end{figure*}

\lstset{
  basicstyle=\ttfamily\footnotesize\linespread{1.08}\selectfont,
  breaklines=true,
  breakatwhitespace=true,
  columns=fullflexible,
  keepspaces=true,
  frame=none,
  breakindent=0.30em,
  breakautoindent=false,
  postbreak=\mbox{},
  aboveskip=0pt,
  belowskip=0pt
}

\begin{figure*}[!p]
\centering
\begin{tcolorbox}[
  enhanced,
  colback=white,
  colframe=black!55,
  boxrule=0.55pt,
  arc=1.2pt,
  left=1.2mm,right=1.2mm,
  top=0.9mm,bottom=0.9mm,
  boxsep=0.4mm
]

\begin{tcolorbox}[
  enhanced,
  colback=white,
  colframe=black!55,
  boxrule=0.45pt,
  arc=1.0pt,
  left=1.4mm,right=1.4mm,
  top=0.7mm,bottom=0.7mm,
  boxsep=0.6mm,
  title={Instruction},
  fonttitle=\bfseries\footnotesize,
  colbacktitle=black!12,
  coltitle=black,
  titlerule=0.45pt
]
\begin{lstlisting}
You are an information extraction assistant...
Please use concise output with no empty fields. The user input is:
\end{lstlisting}
\end{tcolorbox}

\vspace{0.25mm}

\begin{tcolorbox}[
  enhanced,
  colback=white,
  colframe=black!55,
  boxrule=0.45pt,
  arc=1.0pt,
  left=1.4mm,right=1.4mm,
  top=0.7mm,bottom=0.7mm,
  boxsep=0.6mm,
  title={Input Example},
  fonttitle=\bfseries\footnotesize,
  colbacktitle=black!12,
  coltitle=black,
  titlerule=0.45pt
]
\vspace{1.2mm}
\begin{lstlisting}
After 5 days of treatment with IL-2, the patient developed a hemorrhagic lesion that progressed to toxic epidermal necrolysis, as well as grade 4 pancytopenia.
\end{lstlisting}
\end{tcolorbox}

\vspace{0.25mm}

\begin{tcolorbox}[
  enhanced,
  colback=white,
  colframe=black!55,
  boxrule=0.45pt,
  arc=1.0pt,
  left=1.4mm,right=1.4mm,
  top=0.7mm,bottom=0.7mm,
  boxsep=0.6mm,
  title={Output Example},
  fonttitle=\bfseries\footnotesize,
  colbacktitle=black!12,
  coltitle=black,
  titlerule=0.45pt
]
\vspace{0.25mm}
\begin{lstlisting}
{
"adverse event": "developed: Subject: patient; Effect: a hemorrhagic lesion that progressed to toxic epidermal necrolysis, as well as grade 4 pancytopenia; Treatment: 5 days of treatment with IL-2; Treatment.Drug: IL-2; Treatment.Time_elapsed: After 5 days"
}
\end{lstlisting}
\end{tcolorbox}

\end{tcolorbox}
\vspace{-5mm}
\caption{Concise EE output for SA stage.}
\label{fig:prompt-ee-concise}
\end{figure*}

\FloatBarrier
\endgroup

\subsection{Implementation Details}
\label{sec:appendix_settings}

\paragraph{Model and training setup.}
We use Qwen3-4B \cite{yang2025qwen3} as the backbone, implement supervised fine-tuning (CM/SA) with LLaMAFactory~\cite{zheng2024llamafactoryunifiedefficientfinetuning}, and perform GRPO-based optimization (DE) with EasyR1. The experiments were conducted primarily on 8 NVIDIA H20 GPUs.

\paragraph{Stage I: Complete Modeling (CM).}
We conduct supervised fine-tuning in three task-ordered phases.
Phase 1 (NER only) uses 845{,}646 NER instances.
Phase 2 (NER+RE, 2:8) jointly trains on 44{,}951 NER and 179{,}794 RE instances.
Phase 3 (NER+RE+EE, 3:3:4) jointly trains on 7{,}494 NER, 7{,}494 RE, and 9{,}991 EE instances.
We train each CM phase for 2 epochs.

\paragraph{Stage II: Streamlined Alignment (SA).}
We fine-tune on a sampled subset with streamlined targets, consisting of 9{,}000 RE and 9{,}000 EE instances, for 2 epochs.

\paragraph{Stage III: Deep Exploration (DE).}
Starting from the SA checkpoint, we apply GRPO in three task-specific phases.
Phase 1 (NER) uses 80{,}000 instances, with rollout $n{=}4$ per input, and runs for 2 epochs.
Phase 2 (RE) uses 50{,}000 instances, with rollout $n{=}8$ per input, and runs for 3 epochs.
Phase 3 (EE) uses 9{,}991 instances, with rollout $n{=}8$ per input, and runs for 3 epochs.

\paragraph{SFR hyperparameters.}
We instantiate SFR (Section~\ref{subsec:sfr}) with task-specific weights and penalties in Eqs.~\eqref{eq:ner_reward}--\eqref{eq:ee_reward}.

\textbf{NER.}
In Eq.~\eqref{eq:ner_reward}, we set $w_t{=}0.2$, $w_p{=}0.8$, and $\gamma_{\textsc{ner}}{=}1.5$.
For mismatch penalties, we use $\lambda_t{=}0.6$ and $\lambda_p{=}0.2$ on $\Delta(T_g,T_p)$ and $\Delta(P_g,P_p)$, respectively.

\textbf{RE.}
In Eq.~\eqref{eq:re_reward}, we set $w_t{=}0.05$, $w_h{=}0.10$, $w_a{=}0.10$, $w_r{=}0.75$, and $\gamma_{\textsc{re}}{=}1.3$.
For the bounded mismatch penalties, we use $\lambda_t{=}0.15$ and $\lambda_r{=}0.25$ on $D_{\mathrm{jac}}(T_g,T_p)$ and $D_{\mathrm{jac}}(R_g,R_p)$.

\textbf{EE.}
In Eq.~\eqref{eq:ee_reward}, we set $w_E{=}0.05$, $w_T{=}0.15$, $w_F{=}0.8$.
For the coarse-to-fine penalty in Eq.~\eqref{eq:ee_penalty}, we use $\lambda_E{=}1.0$ for event mismatch ($\delta_E$), $\lambda_T{=}0.5$ for trigger mismatch ($\delta_T$), and $\lambda_F{=}0.3$ for residual fine-grained errors $(1-F_{\text{full}})$.

We summarize the stage-wise data mixtures and example counts in Table~\ref{tab:data_mix}, which specifies the task composition for CM/SA and the phase-wise datasets used in DE.

\begin{table}[H]
\centering
\footnotesize
\setlength{\tabcolsep}{5.2pt}
\renewcommand{\arraystretch}{1.12}
\caption{Stage-wise data composition used in ProUIE.}
\label{tab:data_mix}
\begin{tabular*}{\columnwidth}{@{\extracolsep{\fill}} l c c c r}
\toprule
Stage & NER & RE & EE & \#Examples \\
\midrule
CM (Phase 1)            & \checkmark & --         & --         & 845{,}646 \\
CM (Phase 2, 2{:}8)     & \checkmark & \checkmark & --         & 224{,}745 \\
CM (Phase 3, 3{:}3{:}4) & \checkmark & \checkmark & \checkmark & 24{,}979 \\
\midrule
SA (1{:}1)              & -- & \checkmark & \checkmark          & 18{,}000 \\
\midrule
DE (Phase 1)            & \checkmark & --         & --         & 80{,}000 \\
DE (Phase 2)            & --         & \checkmark & --         & 50{,}000 \\
DE (Phase 3)            & --         & --         & \checkmark & 9{,}991 \\
\bottomrule
\end{tabular*}
\end{table}

\lstset{
  basicstyle=\ttfamily\footnotesize\linespread{1.05}\selectfont,
  breaklines=true,
  breakatwhitespace=true,
  columns=fullflexible,
  keepspaces=true,
  frame=none,
  aboveskip=0pt,
  belowskip=0pt
}

\begin{figure}[t]
\centering
\begin{tcolorbox}[
  enhanced,
  colback=white,
  colframe=black!55,
  boxrule=0.55pt,
  arc=1.2pt,
  left=1.2mm,right=1.2mm,
  top=0.9mm,bottom=0.9mm,
  boxsep=0.4mm
]
\begin{tcolorbox}[
  enhanced,
  colback=white,
  colframe=black!55,
  boxrule=0.45pt,
  arc=1.0pt,
  left=1.4mm,right=1.4mm,
  top=0.7mm,bottom=0.7mm,
  boxsep=0.6mm,
  title={Input},
  fonttitle=\bfseries\footnotesize,
  colbacktitle=black!12,
  coltitle=black,
  titlerule=0.45pt
]
\begin{lstlisting}
Labeled data is replaced by a few hand-crafted rules that encode basic syntactic knowledge .
\end{lstlisting}
\end{tcolorbox}

\vspace{0.25mm}

\begin{tcolorbox}[
  enhanced,
  colback=white,
  colframe=black!55,
  boxrule=0.45pt,
  arc=1.0pt,
  left=1.4mm,right=1.4mm,
  top=0.7mm,bottom=0.7mm,
  boxsep=0.6mm,
  title={CM output},
  fonttitle=\bfseries\footnotesize,
  colbacktitle=black!12,
  coltitle=black,
  titlerule=0.45pt
]
\begin{lstlisting}
{
  "conjunction":"",
  "feature of":"",
  "hyponym of":"",
  "used for":"hand-crafted rules, basic syntactic knowledge",
  "part of":"",
  "compare":"",
  "evaluate for":""
}
\end{lstlisting}
\end{tcolorbox}

\vspace{0.25mm}

\begin{tcolorbox}[
  enhanced,
  colback=white,
  colframe=black!55,
  boxrule=0.45pt,
  arc=1.0pt,
  left=1.4mm,right=1.4mm,
  top=0.7mm,bottom=0.7mm,
  boxsep=0.6mm,
  title={CM+SA output},
  fonttitle=\bfseries\footnotesize,
  colbacktitle=black!12,
  coltitle=black,
  titlerule=0.45pt
]
\begin{lstlisting}
{
  "used for":"hand-crafted rules, basic syntactic knowledge"
}
\end{lstlisting}
\end{tcolorbox}

\vspace{0.25mm}

\begin{tcolorbox}[
  enhanced,
  colback=white,
  colframe=black!55,
  boxrule=0.45pt,
  arc=1.0pt,
  left=1.4mm,right=1.4mm,
  top=0.7mm,bottom=0.7mm,
  boxsep=0.6mm,
  title={CM+SA+DE output},
  fonttitle=\bfseries\footnotesize,
  colbacktitle=black!12,
  coltitle=black,
  titlerule=0.45pt
]
\begin{lstlisting}
{
  "used for":"hand-crafted rules, syntactic knowledge"
}
\end{lstlisting}
\end{tcolorbox}

\end{tcolorbox}
\vspace{-3.0mm}
\caption{A \textsc{RE} case study showing progressive refinement from CM to SA to DE.}
\label{fig:case_re}
\end{figure}

\subsection{Case Study Examples}
\label{sec:appendix_case}

We provide a representative \textsc{RE} example to show stage-wise improvements. 
CM predicts the correct relation type but uses a wrong argument span and produces redundant fields; 
SA removes redundant fields but the argument error remains; 
DE recovers the correct argument pair under the same schema.

\subsection{Production-oriented Task Description}
\label{sec:appendix_business}

\paragraph{Business scenario.}
We evaluate ProUIE on a test set from an internal on-device AI search scenario, where the model performs intent understanding for mobile search queries.
The extracted structured results are used to support downstream search and user-facing experiences under a fixed JSON schema.
The service operates at large scale, serving over one hundred million active users.

\paragraph{Slot definitions.}
Table~\ref{tab:slot_definitions} summarizes the meanings of the six representative slots reported in Figure~\ref{fig:nlu}.
All slot values are required to be valid JSON strings that strictly follow the predefined schema.
We report \textit{ExactAcc} at the slot level: a prediction is considered correct if and only if the predicted slot value exactly matches the gold JSON string after normalization (e.g., trimming surrounding whitespace).

\begin{table}[t]
\centering
\small
\setlength{\tabcolsep}{6pt}
\renewcommand{\arraystretch}{1.15}
\caption{Definitions of representative slots in the production-oriented evaluation.}
\label{tab:slot_definitions}
\resizebox{\columnwidth}{!}{
\begin{tabular}{l p{0.78\columnwidth}}
\toprule
\textbf{Slot} & \textbf{Definition} \\
\midrule
body & The main subject of the query (i.e., what the user is searching for or focusing on). \\
action & The action expressed in the query. \\
environment & The environment or scenario described in the query. \\
location & The address/location constraint mentioned in the query. \\
number & The quantity associated with a \texttt{body} in the query. \\
source & The search source indicated in the query (e.g., where to search from). \\
\bottomrule
\end{tabular}}
\end{table}

\end{document}